\documentclass[11pt,a4paper]{article}
\usepackage{hyperref}
\usepackage{acl2018}
\usepackage{times}
\usepackage{latexsym}
\usepackage{booktabs}
\usepackage[nolist]{acronym}
\usepackage{url}
\usepackage{todonotes}
\usepackage{array}
\newcolumntype{L}{>{\arraybackslash}m{13cm}}

\usepackage[inline]{enumitem}
\usepackage{xspace}
\usepackage{listings}
\usepackage{fancyvrb}
\usepackage{comment}
\usepackage{wrapfig}  
\usepackage{caption}
\usepackage{subcaption}
\usepackage{longtable}
\usepackage{multirow}
\usepackage{colortbl}
\usepackage[T1]{fontenc}
\usepackage{tikz}
\usepackage{footnote}
\usepackage{amssymb,amsmath,amsthm}
\usepackage[ruled,vlined,linesnumbered]{algorithm2e}
\usetikzlibrary{shapes,arrows}
\usepackage[figuresright]{rotating}
\usepackage{lineno}

\usepackage[utf8]{inputenc}

\aclfinalcopy 
\def\aclpaperid{XXX} 

\graphicspath{ {img/} }

\definecolor{purple}{rgb}{0.65, 0.12, 0.82}
\definecolor{lightblue}{rgb}{0,0,.7}
\definecolor{orange}{rgb}{1,.7,0}
\definecolor{darkorange}{rgb}{1,.4,0}
\definecolor{darkgreen}{rgb}{.0,.6,.0}
\definecolor{darkblue}{rgb}{0,0,.4}
\definecolor{darkred}{rgb}{.4,0,0}
\definecolor{gray}{rgb}{.3,.3,.3}
\definecolor{darkgray}{rgb}{.4,.4,.4}
\definecolor{shadecolor}{gray}{0.975}
\lstdefinelanguage{SPARQL}{%
	morekeywords=[1]{CONSTRUCT,WHERE,SELECT},
	morekeywords=[2]{AND,FILTER,UNION,OPT,OPTIONAL,MINUS,ORDER,GROUP,BY,DESC,OFFSET,LIMIT},%
	morekeywords=[3]{sameTerm,isBLANK,isLITERAL,isIRI,BOUND,DISTINCT},
	morekeywords=[4]{rdf,rdfs,owl,dbo,res,xsd},
	morekeywords=[5]{>},
	morestring=[b]",%
	alsodigit={-},%
}[keywords,strings]

\newcommand\Bengal{\textsc{Bengal}\xspace}

\title{{BENGAL}: An Automatic Benchmark Generator for Entity Recognition and Linking}

\author{Axel-Cyrille Ngoma Ngomo$^{1,2}$ \, Michael R\"oder$^{1}$ \, Diego Moussallem$^{1,2}$ \, Ricardo Usbeck$^{1}$ \, Ren\'e Speck$^{1,2}$ \\
         $^{1}$Data Science Group, University of Paderborn, Germany \\
         $^{2}$AKSW Research Group, University of Leipzig, Germany \\
         {\tt \{first.lastname\}@upb.de} \\
         {\tt lastname@informatik.uni-leipzig.de}}

\date{}
\begin{acronym}[UML]
	\acro{AOS}{Agricultural Ontology Services}
	\acro{AGRIS}{Agricultural Science and Technology}
	\acro{API}{Application Programming Interface}
	\acro{A2KB}{Annotation to Knowledge Base}
	\acro{BPSO}{Binary Particle-Swarm Optimization}
	\acro{BPMLOD}{Best Practices for Multilingual Linked Open Data}
	\acro{BFS}{Breadth-First-Search}
	\acro{CBD}{Concise Bounded Description}
	\acro{COG}{Content Oriented Guidelines}
	\acro{CSV}{Comma-Separated Values}
	\acro{CCR}{cross-document co-reference}
	\acro{DPSO}{Deterministic Particle-Swarm Optimization}
	\acro{DALY}{Disability Adjusted Life Year}
	\acro{D2KB}{Disambiguation to Knowledge Base}

	\acro{ER}{Entity Resolution}
	\acro{EM}{Expectation Maximization}
	\acro{EL}{Entity Linking}
	\acro{FAO}{Food and Agriculture Organization of the United Nations}
	\acro{GIS}{Geographic Information Systems}
	\acro{GHO}{Global Health Observatory}
	\acro{HDI}{Human Development Index}
	\acro{ICT}{Information and communication technologies}
    \acro{KB}{Knowledge Base}
	\acro{LR}  {Language Resource}
	\acro{LD}  {Linked Data}
	\acro{LLOD}  {Linguistic Linked Open Data}
	\acro{LIMES}{LInk discovery framework for MEtric Spaces}
	\acro{LS}  {Link Specifications}
	\acro{LDIF}{Linked Data Integration Framework}
	\acro{LGD} {LinkedGeoData}
	\acro{LOD} {Linked Open Data}
    \acro{LSTM} {Long-Short Term Memory Layers}
	\acro{MSE}{Mean Squared Error}
	\acro{MWE}{Multiword Expressions}
	\acro{NIF}{NLP Interchange Format}
	\acro{NIF4OGGD}{NLP Interchange Format for Open German Governmental Data}
	\acro{NLP}{Natural Language Processing}
	\acro{NL}{Natural Language}
	\acro{NER}{Named Entity Recognition}
	\acro{NED}{Named Entity Disambiguation}
	\acro{NEL}{Named Entity Linking}
	\acro{NN}{Neural Network}
    \acro{NLG}{Natural Language Generation}
	\acro{OSM}{OpenStreetMap}
	\acro{OWL}{Web Ontology Language}
	\acro{PFM}{Pseudo-F-Measures}
	\acro{PSO}{Particle-Swarm Optimization}
	\acro{QA}{Question Answering}
	\acro{RDF}{Resource Description Framework}
    \acro{REG}{Referring Expression Generation}
	\acro{SKOS}{Simple Knowledge Organization System}
	\acro{SPARQL}{SPARQL Protocol and RDF Query Language}
	\acro{SRL}{Statistical Relational Learning}
	\acro{SF}{surface forms}
    \acro{SW}{Semantic Web}
    \acro{SWT}{Semantic Web Technologies}
	\acro{UML}{Unified Modeling Language}
	\acro{WHO}{World Health Organization}
	\acro{WKT}{Well-Known Text}
	\acro{W3C}{World Wide Web Consortium}
	\acro{YPLL}{Years of Potential Life Lost}
\end{acronym}  
\begin{document}
\maketitle
\begin{abstract}

The manual creation of gold standards for named entity recognition and entity linking is time- and resource-intensive. Moreover, recent works show that such gold standards contain a large proportion of mistakes in addition to being difficult to maintain.
We hence present \Bengal{}, a novel automatic generation of such gold standards 
as a complement to manually created benchmarks.
The main advantage of our benchmarks is that they can be readily generated at any time. They are also cost-effective while being guaranteed to be free of annotation errors.
We compare the performance of 11 tools on benchmarks in English generated by \Bengal{} and on 16 benchmarks created manually. 
We show that our approach can be ported easily across languages by presenting results achieved by 4 tools on both Brazilian Portuguese and Spanish. 
Overall, our results suggest that our automatic benchmark generation approach can create varied benchmarks that have characteristics similar to those of existing benchmarks. Our approach is open-source. 
Our expe-rimental results are available at~\url{http://faturl.com/bengalexpinlg} and the code at \url{https://github.com/dice-group/BENGAL}.
\end{abstract}

\section{Introduction}
\label{intro}

The creating of gold standard is of central importance for the objective assessment and development of approaches all around computer science.  
For example, evaluation campaigns such as BioASQ~\cite{bioasq} have led to an improvement of the F-measure achieved by bio-medical question answering systems by more than 5\%. 
While the manual creation of \ac{NER} and \ac{EL} gold standards (also called benchmarks) has the advantage of yielding resources which reflect human processing, it also exhibits significant disadvantages: 
\begin{enumerate*}[label={\alph*)},font={\color{black}\bfseries}]
\item \emph{Annotation mistakes}: Human annotators have to read through every sentence in the corpus and often (a) miss annotations or (b) assign wrong resources to entities for reasons as various as fatigue or lack of background knowledge (and this even when supported with annotation tools). For example, ~\newcite{eaglet_ESWC} was able to determine that up to 38,453 of the annotations in commonly used benchmarks (see GERBIL~\cite{gerbil} for a list of these benchmarks) were erroneous. A manual evaluation of 25 documents from the ACE2004 benchmark revealed that 195 annotations were missing and 14 of 306 annotations were incorrect. Similar findings were reported for AIDA/CONLL~\cite{conll2003} and OKE2015~\cite{okechallenge}. 
\item  \emph{Volume}: Manually created benchmarks are usually small (commonly $<2,500$ documents, see Table~\ref{tab:features}). Hence, they are of little help when aiming to benchmark the scalability of existing solutions (especially when these solutions use caching).
\item  \emph{Lack of updates}: Manual benchmark generation approaches lead to static corpora which tend not to reflect the newest reference knowledge graphs (also called \ac{KB}s). For example, several of the benchmarks presented in GERBIL~\cite{gerbil} link to outdated versions of Wikipedia or DBpedia. 
\item  \emph{Popularity bias}: \newcite{vanErp2016} show that manual benchmarks are often biased towards popular resources.
\item \emph{Lack of availability}: 
    The lack of benchmarks for resource-poor languages inhibits the development of corresponding \ac{NER} and \ac{EL} solutions.
\end{enumerate*}


Automatic methods are a \emph{viable and supplementary} approach for the generation of gold standards for \ac{NER} and \ac{EL}, especially as they address some of the weaknesses of the manual benchmark creation process. 
\emph{The main contribution of our paper is a novel approach for the automatic generation of benchmarks for \ac{NER} and \ac{EL}} dubbed \Bengal. 
Our approach relies on the abundance of structured data in \ac{RDF} on the Web and is based on \ac{NLG} techniques which verbalize such data to generate automatically annotated natural language statements. 
Our automatic benchmark creation method addresses the drawbacks of manual benchmark generation aforementioned as follows:
\begin{enumerate*}[label={\alph*)},font={\color{black}\bfseries}]
\item It alleviates the human annotation error problem by relying on data in \ac{RDF} which explicitly contain the entities to find. 
\item \Bengal is able to generate arbitrarily large benchmarks. Hence, it can enhance the measurement of both the accuracy and the scalability of approaches.
\item  \Bengal can be updated easily to reflect the newest terminology and reference \ac{KB}s. Hence, it can generate corpora that reflect the newest \ac{KB}s. 
\item  \Bengal is not biased towards popular resources as it can choose entities to include in the benchmark generated following a uniform distribution.
\item \Bengal can be ported to any token-based language. This is exemplified by porting \Bengal to Portuguese and Spanish.
\end{enumerate*}
 

\section{Related Work}
\label{sec:related}


\subsection{Gold Standards for NER and EL}
According to GERBIL~\cite{gerbil}, the 2003 CoNLL shared task~\cite{conll2003} is the most used benchmark dataset for recognition and linking. 
The ACE2004 and MSNBC~\cite{Cucerzan07} news datasets were used by Ratinov et al.~\cite{rat:rot} to evaluate their seminal work on linking to Wikipedia.
Another often-used corpus is AQUAINT, e.g., 
used by Milne and Witten~\cite{milne2008learning}. 
Detailed dataset statistics on some of these benchmarks can be found in Table~\ref{tab:features}. 

A recent uptake of publicly available corpora~\cite{N3,yovisto} based on RDF has led to the creation of many new datasets.
For example, the Spotlight corpus and the KORE 50 dataset were proposed to showcase the usability of RDF-based annotations~\cite{spotlight}. 
The multilingual N3 collection~\cite{N3} was introduced to widen the scope and diversity of NIF-based corpora.
Another recent observation is the shift towards gold standards for micropost documents like tweets. For example, 
the Microposts2014 corpus~\cite{cano2014} was created to  evaluate \ac{NER} on smaller pieces of text. 

Semi-automatic approaches to benchmark creation are commonly crowd-based. 
They use one or more recognizers to create a first set of annotations and then hand over the tasks of refinement and/or linking to crowd workers to improve the quality. Examples of such approaches include \newcite{voyer2010} and CALBC~\cite{rebholz2010calbc}. 
\newcite{OramasASSS16} introduced a voting-based algorithm which analyses the hyperlinks presented in the input texts retrieved from different disambiguation systems such as Babelfy~\cite{moro2014multilingual}. Each entity mention in the input text is linked based on the degree of agreement across three \ac{EL} systems.


\Bengal is the first automatic approach that makes use of structured data and can be replicated on any RDF KB for \ac{EL} benchmarks. 


\subsection{NLG for the Web of Data}
A plethora of works have investigated the generation of \ac{NL} texts from \ac{SWT} such as ~\newcite{staykova2014natural,bouayad2014natural}. However, the generation of NL from RDF has only recently gained momentum. This attention comes from the great number of published works such as~\cite{cimiano2013exploiting,duma2013generating,ell2014language,biran2015discourse} which used \ac{RDF} as an input data and achieved promising results. 
Moreover, the works published in the WebNLG \cite{colin2016webnlg} challenge, which used deep learning techniques such as \cite{sleimi2016generating,mrabet2016aligning}, also contributed to this interest. 
\ac{RDF} has also been showing promising benefits to the generation of benchmarks for evaluating \ac{NLG} systems, e.g.,~\cite{gardent2017creating,perez2016building,mohammed2016category,schwitter2004controlled,hewlett2005effective,sun2006domain}. However, RDF has never been used for creating NER and NEL benchmarks. \Bengal{} addresses this research gap. 

\section{The \Bengal approach}
\label{sec:bengal}
\begin{figure*}[ht]
\centering
    \includegraphics[width=\textwidth]{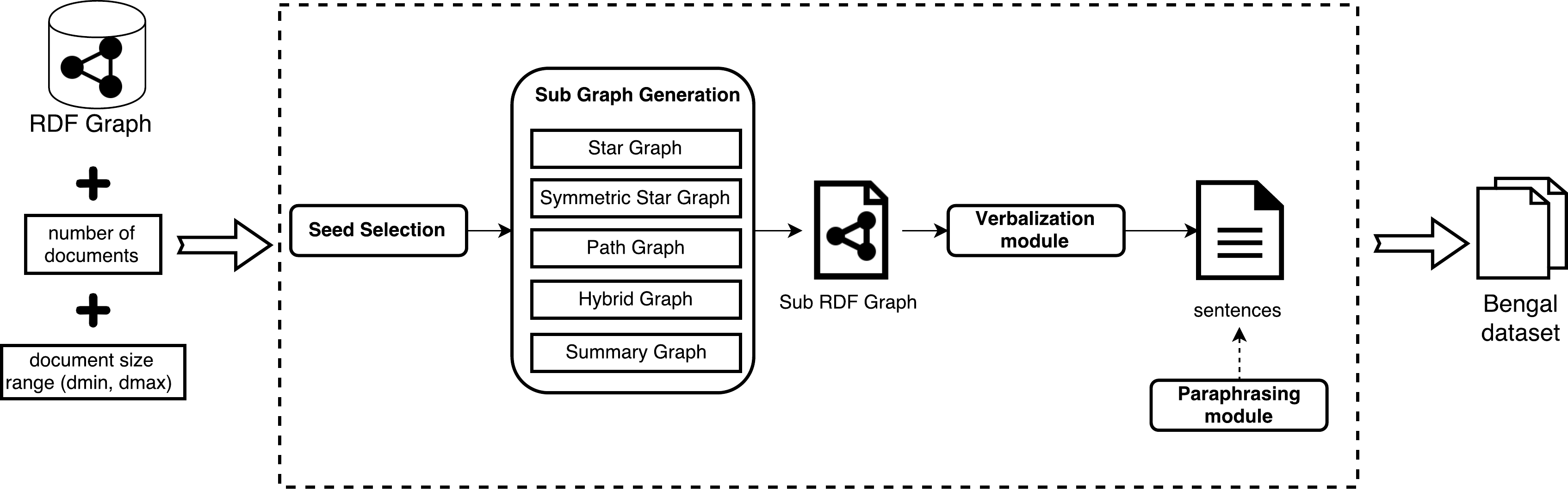}
    \caption{Overview of the \Bengal approach.}
    \label{fig:arch}
\end{figure*}

\Bengal is based on the observation that more than 150 billion facts pertaining to more than 3 billion entities are available in machine-readable form on the Web (i.e., as RDF triples).\footnote{\url{http://stats.lod2.eu}}
The basic intuition behind our approach is hence as follows: \emph{Given that \ac{NER} and \ac{EL} are often used in pipelines for the extraction of machine-readable facts from text, we can invert the pipeline and go from facts to text}, thereby using the information in the facts to produce a gold standard that is \emph{guaranteed to contain no errors}. 
In the following, we begin by giving a brief formal overview of \ac{RDF}.
Thereafter, we present how we use \ac{RDF} to generate \ac{NER} and \ac{EL} benchmarks automatically and at scale. 

\subsection{Preliminaries and Notation}
\label{subsec:notation}
\subsubsection{RDF}
The notation presented herein is based on the RDF 1.1 specification.
An RDF graph $G$ is a set of facts. 
Each fact is a triple $t = (s, p, o) \in (R \cup B) \times P \times (R \cup B \cup L)$ where $R$ is the set of all resources (i.e., things of the real world), $P$ is the set of all predicates (binary relations), $B$ is the set of all blank nodes (which basically express existential quantification) and $L$ is the set of all literals (i.e., of datatype values). We call the set $R \cup P \cup L \cup B$ our universe and call its elements entities.
A fragment of DBpedia\footnote{\url{http://dbpedia.org}} is shown below. We will use this fragment in our examples. For the sake of space, our examples are in English. However, note that we ported \Bengal to Portuguese and Spanish so as to exemplify that it is not biased towards a particular language. Also, the morphological richness of both led us to choose them as languages.

 
\begin{lstlisting}[caption=Example RDF dataset.,label=lst:example,language=SPARQL]
:Albert_Einstein dbo:birthPlace :Ulm . 	
:Albert_Einstein dbo:deathPlace :Princeton . 	
:Albert_Einstein rdf:type dbo:Scientist . 	
:Albert_Einstein dbo:field :Physics .		
:Ulm dbo:country :Germany.						
:Albert_Einstein rdfs:label "Albert Einstein"@en. 
\end{lstlisting}

\subsubsection{Benchmarks}
We define a benchmark as a set $C$ of annotated documents $D_i$. 
Each document $D_i$ is a sequence of characters $s_{i1} \ldots s_{in}$. Each subsequence $s_{ij} \ldots s_{ik}$ (with $j < k$) of the document $D_i$ which stands for a resource $r \in R$ is assumed to be marked as such. We model the marking of resources by the function $m: C \times \mathbb{N} \times \mathbb{N} \rightarrow R$ and write $m(D_i, j, k) = r$ to signify that the substring $s_{ij} \ldots s_{ik}$ stands for the resource $r$. In case the substring $s_{ij} \ldots s_{ik}$ does not stand for a resource, we write $m(D_i,j,k) = \epsilon$. Let $D_0$ be the example shown in Listing~\ref{lst:text}. We would write $m(D_0, 0, 14) = \texttt{:AlbertEinstein}$. 

\begin{lstlisting}[caption=Example sentence.,label=lst:text]
Albert Einstein was born in Ulm.
\end{lstlisting}

\subsection{Verbalization}
The notation and formal framework for verbalization in \Bengal are based on SPARQL2NL~\cite{ngo+13a}. 
Let $W$ be the set of all words in the dictionary of our target language (e.g., English). We define the realization function $\rho: R \cup P \cup L \rightarrow W^*$ as the function which maps each entity to a word or sequence of words from the dictionary. 
Formally, the goal of our NLG approach is to devise an extension of $\rho$ to conjunctions of RDF triples.
This extension maps all triples $t$ to their realization $\rho(t)$ and defines how these atomic realizations are to be combined.
We denote the extension of $\rho$ by the same label $\rho$ for the sake of simplicity.
We adopt a rule-based approach to devise the extension of $\rho$, where the rules extending $\rho$ to RDF triples are expressed in a conjunctive manner. This means that for premises $P_1,\ldots,P_n$ and consequences $K_1,\ldots,K_m$ we write $P_1 \wedge \ldots \wedge P_n \Rightarrow K_1 \wedge \ldots \wedge K_m$. 
The premises and consequences are explicated by using an extension of the Stanford dependencies.\footnote{For a complete description of the vocabulary, see \url{http://nlp.stanford.edu/software/dependencies_manual.pdf}.} 
We rely especially on the constructs explained in Table~\ref{tab:dependencies}.
For example, a possessive dependency between two phrase elements $e_1$ and $e_2$ is represented as $\texttt{poss}(e_1,e_2)$.
For the sake of simplicity, we sometimes reduce the construct $\texttt{subj(y,x)}\wedge \texttt{dobj(y,z)}$ to the triple \texttt{(x,y,z)} $\in W^3$.

 \begin{table*}[htb]
 	\centering
 	\small
 	\caption{Dependencies used by \Bengal.}
 	\label{tab:dependencies}
 	\begin{tabular}{@{}lp{13.5cm}@{}}
 		\toprule
 		\textbf{Dependency}		&	\textbf{Explanation}	\\
 		\midrule
 		\texttt{cc} & Stands for the relation between a conjunct and a given conjunction (in most cases {\sf and} or {\sf or}). For example in the sentence {\sf John eats an apple and a pear}, \texttt{cc(PEAR,AND)} holds. We mainly use this construct to specify reduction and replacement rules. \\
 		\texttt{conj}$^*$ &	Used to build the \emph{conjunction} of two phrase elements, e.g. \texttt{conj(subj(EAT,JOHN), subj(DRINK,MARY))} stands for {\sf John eats and Mary drinks}. 
 		\texttt{conj} is not to be confused with the logical conjunction $\wedge$, which we use to state that two dependencies hold in the same sentence. 
 		For example \texttt{subj(EAT,JOHN) $\wedge$ dobj(EAT,FISH)} is to be read as {\sf John eats fish}.\\
 		\texttt{dobj}	& Dependency between a verb and its \emph{direct object}, for example \texttt{dobj(EAT,APPLE)} expresses {\sf to eat an/the apple}.\\
 		\texttt{nn} &	The \emph{noun compound modifier} is used to modify a head noun by the means of another noun. For instance \texttt{nn(FARMER,JOHN)} stands for {\sf farmer John}.\\
 		\texttt{poss} &	Expresses a possessive dependency between two lexical items, for example \texttt{poss(JOHN,DOG)} expresses {\sf John's dog}.\\
 		\texttt{subj}	& Relation between \emph{subject} and verb, for example \texttt{subj(BE,JOHN)} expresses {\sf John is}.\\
 		\bottomrule
 	\end{tabular}
 \end{table*}

\subsection{Approach}

\Bengal assumes that it is given 
(1) an RDF graph $G \subseteq  (R \cup B) \times P \times (R \cup B \cup L)$, 
(2) a number of documents to generate, 
(3) a minimal resp. maximal document size (i.e., number of triples to use during the generation process) $d_{min}$ resp. $d_{max}$,  
(4) a set of restrictions pertaining to the resources to generate and 
(5) a strategy for generating single documents. 
Given the graph $G$, \Bengal begins by selecting a set of \emph{seed resources} from $G$ based on the restrictions set using parameter (4). 
Thereafter, it uses the strategy defined via parameter (5) to select a subgraph of $G$. This subgraph contains a randomly selected number $d$ of triples with $d_{min} \leq d \leq d_{max}$. The subgraph is then verbalized. The verbalization is annotated automatically and finally returned as a single document. Each single document then may be paraphrased if this option is chosen in the initial phase. This process is repeated as many times as necessary to reach the predefined number of documents. 
In the following, we present the details of each step underlying our benchmark generation process displayed in \autoref{fig:arch}. 

\subsubsection{Seed Selection}

Given that we rely on RDF, we model the seed selection by means of a SPARQL SELECT query with one projection variable. Note that we can use the wealth of SPARQL to devise seed selection strategies of arbitrary complexity. However, given that NER and EL frameworks commonly focus on particular classes of resources, we are confronted with the condition that the seeds must be instances of a set of classes, e.g., \texttt{:Person}, \texttt{:Organization} or \texttt{:Place}. The SPARQL query for our example dataset would be as follows:

\begin{lstlisting}[caption={Example seed selection query.},label={lst:seeds},language=SPARQL]
SELECT ?x WHERE { {?x a :Person.} UNION {?x a :Organization.} UNION {?x a :Place.} } 
\end{lstlisting}

\subsubsection{Subgraph Generation}
Our approach to generating subgraphs is reminiscent of SPARQL query topologies as available in SPARQL query benchmarks. 
As these queries (e.g., FEASIBLE\footnote{\url{http://aksw.org/Projects/Feasible}} queries) describe real information needs, their topology must stand for the type of information that is necessitated by applications and humans. We thus distinguish between three main types of subgraphs to be generated from RDF data: 
(1) \emph{star graphs} provide information about a particular entity (e.g, the short biography of a person);
(2) \emph{path graphs} describe the relations between two entities (e.g., the relation between a gene and a side-effect);
(3) \emph{hybrid graphs} are a mix of both and commonly describe a specialized subject matter involving several actors (e.g., a description of the cast of a movie).

\textit{Star Graphs.}
For each $s_i \in S$, we gather all triples of the form $t = (s_i, p, o) \in R \times P \times (R \cup L)$.\footnote{Note that we do not consider blank nodes as they cannot be verbalized due to the existential quantification they stand for.} The triples are then added to a list $L(s_i)$ sorted in descending order according to a hash function $h$. After randomly selecting a document size $d$ between $d_{min}$ and $d_{max}$, we select $d$ random triples from $L(s_i)$. For the dataset shown in Listing \ref{lst:example} and $d=2$, we would for example get Listing~\ref{lst:star}.
\begin{lstlisting}[caption=Example dataset generated by the star strategy.,label=lst:star]
:AlbertEinstein :birthPlace :Ulm . 	
:AlbertEinstein :deathPlace :Princeton . 	
\end{lstlisting}

\textit{Symmetric Star Graphs.} As above with $t \in \{(s_i, p, o) \in G \vee (o, p, s_i) \in G$\}.

\textit{Path Graphs.} For each $s_i \in S$, we begin by computing list $L(s_i)$ as in the symmetric star graph generation. Then, we pick a random triple $(s_i, p, o)$ or $(o, p, s_i)$ from $L(s_i)$ that is such that $o$ is a resource. We then use $o$ as seed and repeat the operation until we have generated $d$ triples, where $d$ is randomly generated as above. For the example dataset shown in Listing \ref{lst:example} and $d=2$, we would for example get Listing~\ref{lst:path}. 
\begin{lstlisting}[caption=Example dataset generated by the path strategy.,label=lst:path]
:AlbertEinstein :birthPlace :Ulm . 	
:Ulm :country :Germany . 	
\end{lstlisting}

\textit{Hybrid Graphs.} This is a 50/50-mix of the star and path graph generation approaches. In each iteration, we choose and apply one of the two strategies above randomly. For example, the hybrid graph generation can generate:

\begin{lstlisting}[caption=Example dataset generated by the hybrid strategy.,label=lst:hybrid]
:AlbertEinstein :birthPlace :Ulm . 	
:AlbertEinstein :deathPlace :Princeton .
:Ulm :country :Germany . 	
\end{lstlisting}

\textit{Summary Graph Generation.} This last strategy is a specialization of the star graph generation where the set of triples to a resource is not chosen randomly. 
Instead, for each class (e.g., \texttt{:Person}) of the input KB, we begin by filtering the set of properties and only consider properties that (1) have the said class as domain and (2) achieve a coverage above a user-set threshold (60\% in our experiments) (e.g., \texttt{:birthPlace}, \texttt{:deathPlace}, \texttt{:spouse}). 
We then build a property co-occurence graph for the said class in which the nodes are the properties selected in the preceding step and the co-occurence of two properties $p_1$ and $p_2$ is the instance $r$ of the input class where  
$\exists o_1, o_2: (r, p_1, o_1) \in K \wedge (r, p_2, o_2) \in K$. The resulting graph is then clustered (e.g., by using the approach presented by~\newcite{ngomo2009borderflow}). We finally select the clusters which contain the properties with the highest frequencies in $K$ that allow the selection of at least $d$ triples from $K$. For example, if \texttt{:birthPlace} (frequency = 10), \texttt{:deathPlace} (frequency = 10) were in the same cluster while \texttt{:spouse} (frequency = 8) were in its own cluster, we would choose the pair (\texttt{:birthPlace}, \texttt{:deathPlace}) and return the corresponding triples for our input resource. Hence, we would return Listing~\ref{lst:star} for our running example.

\subsubsection{Verbalization module}
The verbalization (micro-planning) strategy for the first four strategies consists of verbalizing each triple as a single sentence and is derived from~SPARQL2NL~\cite{ngo+13a}. 
To verbalize the subject of the triple $t=(s, p, o)$, we use one of its labels according to \newcite{ell2011labels} (e.g., the \texttt{rdfs:label}). 
If the object $o$ is a resource, we follow the same approach as for the subject. 
Importantly, the verbalization of a triple $t = (s, p, o)$ depends mostly on the verbalization of the predicate \texttt{p} (see Table \ref{tab:dependencies} for semantics).
If \texttt{p} can be realized as a noun phrase, then a possessive clause can be used to express the semantics of $(s, p, o)$. For example, if $p$ can be verbalized as a nominal compound like \texttt{birth place}, then the verbalization $\rho(s, p, o)$ of the triple is as follows:  \texttt{poss}($\rho(p)$,$\rho(s)$) $\wedge$ \texttt{subj}(\text{\texttt{BE}},$\rho(p)$) $\wedge$  \texttt{dobj}(\text{\texttt{BE}},$\rho(o)$).
In case \texttt{p}'s realization is a verb, then the triple can be verbalized as \texttt{subj}($\rho(p)$,$\rho(s)$) $\wedge$ \texttt{dobj}($\rho(p)$,$\rho(o)$).
In our example, verbalizing (\texttt{:AlbertEinstein}, \texttt{dbo:birthPlace}, \texttt{:Ulm}) would thus lead to \texttt{Albert Einstein's birth place is Ulm.}, 
as \texttt{birth place} is a noun.

In the case of summary graphs, we go beyond the verbalization of single sentences and merge sentences that were derived from the same cluster. For example, if $p_1$ and $p_2$ can be verbalized as nouns, then we apply the following rule:
$\rho(s, p_1, o_1)\land\rho(s, p_2, o_2)  \Rightarrow  \text{\texttt{conj}}(\text{\texttt{poss}}(\rho(p_1),\rho(s)) 
 \land \text{\texttt{subj}}(\text{\texttt{BE}}_1,\rho(p_1)) \land \text{\texttt{dobj}}(\text{\texttt{BE}}_1,\rho(o_1))  \nonumber
 \land \text{\texttt{poss}}(\rho(p_2),\rho(\text{\texttt{pronoun}}(s)))\nonumber \land \text{\texttt{subj}}(\text{\texttt{BE}}_2,\rho(p_2)) \land \text{\texttt{dobj}}(\text{\texttt{BE}}_2,\rho(o_2))\nonumber$.
Note that \texttt{pronoun(s)} returns the correct pronoun for a resource based on its type and gender. Therewith, we can generate \texttt{Albert Einstein's birth place is Ulm and his death pla\-ce is Prin\-ce\-ton}.

\subsubsection{Paraphrasing}
With this step, \Bengal avoids the generation of a large number of sentences that share the same terms and the same structure. 
Additionally, this step makes the use of reverse engineering strategies for the generation more difficult as it increases the diversity of the text in the benchmarks.
Our paraphrasing is largely based on~\newcite{androutsopoulos2010survey} and runs as follows:
\begin{enumerate}
    \item Change the structure of the sentence: We use the location of verbs in each sentence to randomly change passive into active structures and vice-versa. Sentences which describe type information (e.g., \texttt{Einstein is a person}) are not altered. 
    \item Replace synonyms: We use POS tags to select alternative labels from the knowledge base and a reference dictionary to replace entity labels by a synonym.
\end{enumerate} 
An example of a paraphrase generated by \Bengal is shown in Listing \ref{lst:parasummary}.

\begin{lstlisting}[caption=Example Paraphasing at Summary Generation,label=lst:parasummary]
(*\bfseries Original:*) Edmund Pettus Bridge is a bridge. It crosses Alabama River. Its type is Through arch bridge. It was declared a National Historic Landmark on March 11, 2013.

(*\bfseries Paraphrased:*) Edmund Pettus Bridge is a bridge. It crosses Alabama River. Through arch bridge is its type. Pettus was declared a National Historic Landmark on March 11, 2013.
\end{lstlisting}

\section{Experiments and Results}
\label{sec:eval}

We generated 13 datasets in English (B1-B13), 4 datasets in Brazilian Portuguese and 4 datasets in Spanish to evaluate our approach.\footnote{All \Bengal datasets can be found at \url{https://hobbitdata.informatik.uni-leipzig.de/bengal/}} 
B1 to B10 were generated by running our five sub-graph generation methods with and without paraphrasing.
The number of documents was set to 100 while $(d_{min}, d_{max})$ was set to $(1,5)$.
B11 shows how \Bengal can be used to evaluate the scalability of approaches.\footnote{The scalability results are available at \url{https://goo.gl/9mnbwC} and cannot be presented herein due to space limitations.} 
Here, we used the hybrid generation strategy to generate 10,000 documents.
B12 and B13 comprise 10 longer documents each with $d_{min}$ set to 90.
For B12, we focused on generating a high number of entities in the documents while B13 contains less entities but the same number of documents.

We compared B1-B13 with the 16 manually created gold standards for English found in GERBIL.
The comparison was carried out in two ways.
First, we assessed the features of the datasets.
Then, we compared the micro F-measure of 11 NER and EL frameworks on the manually and automatically generated datasets. 
We chose to use these 11 frameworks because they are included in GERBIL.
This inclusion ensures that their interfaces are compatible and their results comparable. 
In addition, we assessed the performance of multi-lingual NER and EL systems on the datasets P1-P4 to show that \Bengal can be easily ported to languages other than English.

\subsection{English Dataset features}
The first aim of our evaluation was to quantify the variability of the datasets B1--B13 generated by \Bengal. 
To this end, we compared the distribution of the part of speech (POS) tags of the \Bengal datasets with those of the 16 benchmark datasets. 
An analysis of the Pearson correlation of these distributions revealed that the manually created datasets (D1--D16) have a high correlation (0.88 on average) with a minimum of 0.61 (D10--D16).
The correlation of the POS tag distributions between \Bengal datasets and a manually created dataset vary between 0.34 (D7--B11) and  0.89 (D14--B9) with an average of 0.67.
This shows that \Bengal datasets can be generated to be similar to manually created datasets (D14--B9) as well as to be very different to them (D7--B11). Hence, \Bengal can be used for testing sentence structures that are not common in the current manually generated benchmarks.\footnote{Our complete results at \url{https://goo.gl/TBDxCa}.}

We also studied the distribution of entities and tokens across the datasets in our evaluation. 
Table~\ref{tab:features} gives an overview of these distributions, where $E$ is the set of entities in the corpus $C$.
The distribution of values for the different features is very diverse across the different manually created datasets. 
This is mainly due to (1) different ways to annotate entities and (2) the domains of the datasets (news, description of entities, microposts).
As shown in Table \ref{tab:features}, 
\Bengal can be easily configured to generate a wide variety of datasets with similar quality and number of documents to those of real datasets. This is mainly due to our approach being able to generate benchmarks ranging from (1) benchmarks with sentences containing a large number of entities without any filler terms (high entity density) to (2) benchmarks which contain more information pertaining to entity types and literals (low entity density).



\begin{table*}[h!tb]
\centering
\caption{Excerpt of the features of the datasets used in our evaluation. The datasets B4, B6, B8 and B10 are paraphrased versions of B3, B5, B7 resp. B9 and share similar characteristics.}
\footnotesize
\label{tab:features}
\begin{tabular}{@{}llrrrrrr@{}}
\toprule
ID & Name & Doc. $|C|$ & Tokens $|T|$ & Entities $|E|$& $|T|/|C|$ & $|E|/|C|$  & $|E|/|T|$ \\
\midrule
D1 & ACE2004 & 57 & 21312 & 306 & 373.9 & 5.4 & 0.01 \\
D2 & AIDA/CoNLL-Complete & 1393 & 245008 & 34929 & 175.9 & 25.1 & 0.14 \\
D8 & IITB & 104 & 66531 & 18308 & 639.7 & 176.0 & 0.28 \\
D11 & Microposts2014-Train & 2340 & 40684 & 3822 & 17.4 & 1.6 & 0.09 \\
D15 & OKE 2015 Task 1 evaluation & 101 & 3064 & 664 & 30.3 & 6.6 & 0.22 \\
\midrule
B1 & \Bengal Path 100 & 100 & 1202 & 362 & 12.02 & 3.6 & 0.30 \\
B2 & \Bengal Path Para 100 & 100 & 1250 & 362 & 12.5 & 3.6 & 0.29 \\
B3 & \Bengal Star 100 & 100 & 3039 & 880 & 30.39 & 8.8 & 0.29 \\
B5 & \Bengal Sym 100 & 100 & 2718 & 725 & 27.18 & 7.25 & 0.26 \\
B9 & \Bengal Summary 100 & 100 & 2033 & 637 & 20.33 & 6.37 & 0.31 \\
B11 & \Bengal Hybrid 10000 & 10000 & 556483 & 165254 & 55.6 & 16.5 & 0.30 \\
B12 & \Bengal Hybrid Long 10 & 10 & 9162 & 2417 & 241.7 & 916.2 & 0.26 \\
B13 & \Bengal Star Long 10 & 10 & 7369 & 316 & 31.6 & 736.9 & 0.04 \\
\bottomrule
\end{tabular}
\end{table*}

\subsection{Annotator performance}

\begin{table*}[h!tb]
    \centering
    \footnotesize{
     \captionof{table}{Excerpt of micro F1-scores of the annotators for the A2KB  experiments on chosen datasets. N/A means that the annotator stopped with an error.}
    \label{tab:f1-scores}
    \begin{tabular}{@{}llrrrrrrrrrrrr@{}}
    \toprule
    \multicolumn{1}{c}{\rotatebox{90}{Experiment}} & \multicolumn{1}{c}{\rotatebox{90}{Dataset ID}} & \multicolumn{1}{c}{\rotatebox{90}{AIDA}} & \multicolumn{1}{c}{\rotatebox{90}{Babelfy}} & \multicolumn{1}{c}{\rotatebox{90}{Spotlight}} & \multicolumn{1}{c}{\rotatebox{90}{Dexter}} & \multicolumn{1}{c}{\rotatebox{90}{E.eu}} & \multicolumn{1}{c}{\rotatebox{90}{FOX}} &
    \multicolumn{1}{c}{\rotatebox{90}{FRED}} & \multicolumn{1}{c}{\rotatebox{90}{FREME}} & \multicolumn{1}{c}{\rotatebox{90}{WAT}} &
    \multicolumn{1}{c}{\rotatebox{90}{xLisa-NER}} &
    \multicolumn{1}{c}{\rotatebox{90}{xLisa-NGRAM}} &\\
    \midrule
    \multirow{10}{*}{\rotatebox{90}{A2KB}}
    & D1 & 0.26 & 0.13 & 0.18 & 0.21 & 0.14 & 0.13 & N/A & 0.19 & 0.25 & 0.36 & 0.27 \\
    & D2 & 0.68 & 0.45 & 0.54 & 0.47 & 0.39 & 0.51 & N/A & 0.34 & 0.67 & 0.43 & 0.36 \\
    & D8 & 0.14 & 0.13 & 0.26 & 0.21 & 0.15 & 0.10 & N/A & 0.07 & 0.14 & 0.07 & 0.23 \\
    & D11 & 0.38 & 0.31 & 0.45 & 0.39 & 0.36 & 0.32 & 0.07 & 0.25 & 0.40 & 0.36 & 0.32 \\
    & D15 & 0.57 & 0.41 & 0.46 & 0.47 & 0.28 & 0.55 & 0.33 & 0.27 & 0.53 & 0.53 & 0.47 \\
    \cmidrule{2-13}
    & B1 & 0.65 & 0.47 & 0.69 & 0.70 & 0.39 & 0.50 & 0.45 & 0.49 & 0.61 & 0.45 & 0.61 \\
    & B2 & 0.67 & 0.49 & 0.68 & 0.70 & 0.38 & 0.54 & 0.41 & 0.47 & 0.61 & 0.44 & 0.62 \\
    & B3 & 0.62 & 0.48 & 0.57 & 0.65 & 0.27 & 0.47 & 0.35 & 0.38 & 0.53 & 0.36 & 0.43 \\
    & B5 & 0.42 & 0.40 & 0.42 & 0.44 & 0.17 & 0.34 & 0.29 & 0.30 & 0.35 & 0.24 & 0.33 \\
    & B9 & 0.51 & 0.39 & 0.57 & 0.52 & 0.26 & 0.43 & 0.39 & 0.30 & 0.46 & 0.44 & 0.51 \\
    & B11 & 0.68 & 0.68 & 0.69 & 0.74 & 0.24 & 0.49 & 0.41 & 0.47 & 0.65 & 0.44 & 0.51 \\
    & B12 & 0.83 & N/A & 0.79 & 0.84 & 0.40 & 0.73 & N/A & 0.50 & 0.79 & 0.23 & 0.28 \\
    & B13 & 0.33 & 0.38 & 0.33 & 0.40 & 0.11 & 0.17 & N/A & 0.22 & 0.45 & 0.44 & 0.50 \\
    \bottomrule
    \end{tabular}
   }
\end{table*}
We used GERBIL to evaluate the performance of 11 annotators on the manually created as well as the \Bengal datasets.
We evaluated the annotators within an A2KB (annotation to knowledge base) experiment setting:
Each document of the corpora was sent to each annotator.
The annotator had to find and link all entities to a reference KB (here DBpedia).
We measured both the performance of the NER and the EL steps. 

Table~\ref{tab:f1-scores} shows the micro F1-score of the different annotators on chosen datasets.
The manually created datasets showed diverse results.
We analyzed the results further by using the F1-scores of the annotators as features of the datasets.
Based on these feature vectors, we calculated the Pearson correlations between the datasets to identify datasets with similar characteristics.\footnote{All values are at \url{http://goo.gl/Mg3rE1}.}
The Pearson correlations of the F-measures achieved by the different annotators 
on the AIDA/CoNLL datasets (D2--D5) are very high (0.95--1.00) while the correlation between the results on the Spotlight corpus (D7) and N3-Reuters-128 (D13) is around -0.62.
The results on D1 and D12--D15 have a correlation to the AIDA/CoNLL results (D2--D5) that is higher than 0.5.
In contrast, the correlations of D7 and D8 to the AIDA/CoNLL datasets range from -0.54 to -0.36.
These correlations highlight the diversity of the manually created datasets and suggest that creating an approach which emulates all datasets is non-trivial.

Like the correlations between the manually created datasets,  the correlations between the results achieved on \Bengal datasets and hand-crafted datasets vary.
The results on \Bengal correlate most with the results on the OKE 2015 data.  
The highest correlations were achieved with the OKE 2015 Task 1 dataset and 
range between 0.89 and 0.92.
This suggests that our benchmark can emulate entity-centric benchmarks.
The correlation of \Bengal with OKE is however reduced to 0.82 in D13, suggesting that \Bengal can be parametrized so as to diverge from such benchmarks. A similar observation can be made for the correlation D12 and ACE2004, where the correlation increased with the size of the documents in the benchmark. The correlation between the results across \Bengal datasets varies between 0.54 and 1, which further supports that \Bengal can generate a wide range of diverse datasets.

\subsection{Annotator Performance on Spanish and Brazilian Portuguese}

We implemented \Bengal for Brazilian Portuguese by using the RDF verbalizer presented in~\newcite{LRECMoussallem2017} and ran four multilingual \ac{NER} and \ac{EL} (MAG~\cite{Moussallem2017}, DBpedia Spotlight, Babelfy, and PBOH~\cite{Ganea:2016:PBM:2872427.2882988}) frameworks thereon. We also evaluated the performance of these annotators on subsets of the HAREM datasets~\cite{freitas2010second}\footnote{All Portuguese results at \url{http://faturl.com/bengalpt}.}. We then extended this verbalizer to Spanish using the adaption of SimpleNLG to Spanish~\cite{soto2017adapting}. We generated Spanish \Bengal datasets and evaluated the aforementioned \ac{NER} and \ac{EL} systems on them. \footnote{All Spanish results at \url{http://faturl.com/bengales}.} We also included VoxEL~\cite{voxeliswc2018}, a recent gold standard for Spanish. 
While the extension of \Bengal to Portuguese is an important result in itself, our results also provide additional insights in the \ac{NER} and \ac{EL} performance of existing solutions. Our results suggest that existing solutions are mostly biased towards a high precision but often achieve a lower recall on this language. For example, both Spotlight's and Babelfy's recall remain below 0.6 in most cases while their precision goes up to 0.9. This clearly results from the lack of training data for these resource-poor languages. In contrast, 
the Spanish annotators presented low but consistent results, which  confirms the lack of training data of these approaches on Spanish.

\section{Discussion and Conclusion}
\label{sec:discussion}
We presented and evaluated \Bengal, an approach for the automatic generation of NER and EL benchmarks. Our results suggest that our approach can generate diverse benchmarks with characteristics similar to those of a large proportion of existing benchmarks in several languages. 

Overall, our results suggest that \Bengal benchmarks can ease the development of NER and EL tools (especially for resource-poor languages) by providing developers with insights into their performance at virtually no cost. Hence, \Bengal can  improve the push towards better NER and EL frameworks. In future work, we plan to extend the ability of \Bengal to generate longer and more complex sentences as well as the capability of generating different surface forms for a given entity by relying on referring expression models such as NeuralREG model~\cite{thiagoacl2018}. We also intend to provide thorough evaluations of annotators across other resource-poor languages and create corresponding datasets to push the development of tools to process these languages.




\section*{Acknowledgements} This work has been supported by the H2020 project HOBBIT (GA no. 688227) as well as the BMVI projects LIMBO (project no. 19F2029C), OPAL (project no. 19F20284) and also supported by the German Federal Ministry of Education and Research (BMBF) within ’KMU-innovativ: Forschung für die zivile Sicherheit’ in particular ’Forschung für die zivile Sicherheit’ and the project SOLIDE (no. 13N14456). The authors gratefully acknowledge financial support from the German Federal Ministry of Education and Research within Eurostars, a joint programme of EUREKA and the European Community under the project E! 9367 DIESEL and E! 9725 QAMEL.

\bibliography{inlg2018}

\begin{thebibliography}{43}
\expandafter\ifx\csname natexlab\endcsname\relax\def\natexlab#1{#1}\fi

\bibitem[{Androutsopoulos and Malakasiotis(2010)}]{androutsopoulos2010survey}
Ion Androutsopoulos and Prodromos Malakasiotis. 2010.
\newblock A survey of paraphrasing and textual entailment methods.
\newblock \emph{Journal of Artificial Intelligence Research}, pages 135--187.

\bibitem[{Biran and McKeown(2015)}]{biran2015discourse}
Or~Biran and Kathleen McKeown. 2015.
\newblock Discourse planning with an n-gram model of relations.
\newblock In \emph{EMNLP}, pages 1973--1977.

\bibitem[{Bouayad-Agha et~al.(2014)Bouayad-Agha, Casamayor, and
  Wanner}]{bouayad2014natural}
Nadjet Bouayad-Agha, Gerard Casamayor, and Leo Wanner. 2014.
\newblock Natural language generation in the context of the semantic web.
\newblock \emph{Semantic Web}, 5(6):493--513.

\bibitem[{{Cano Basave} et~al.(2014){Cano Basave}, Rizzo, Varga, Rowe,
  Stankovic, and Dadzie}]{cano2014}
Amparo~Elizabeth {Cano Basave}, Giuseppe Rizzo, Andrea Varga, Matthew Rowe,
  Milan Stankovic, and Aba-Sah Dadzie. 2014.
\newblock Making sense of microposts ({\#microposts2014}) named entity
  extraction \& linking challenge.
\newblock In \emph{Proceedings of 4th Workshop on Making Sense of Microposts}.

\bibitem[{Castro~Ferreira et~al.(2018)Castro~Ferreira, Moussallem,
  K{\'a}d{\'a}r, Wubben, and Krahmer}]{thiagoacl2018}
Thiago Castro~Ferreira, Diego Moussallem, {\'A}kos K{\'a}d{\'a}r, Sander
  Wubben, and Emiel Krahmer. 2018.
\newblock \href {http://aclweb.org/anthology/P18-1182} {Neuralreg: An
  end-to-end approach to referring expression generation}.
\newblock In \emph{Proceedings of the 56th Annual Meeting of the Association
  for Computational Linguistics (Volume 1: Long Papers)}, pages 1959--1969.
  Association for Computational Linguistics.

\bibitem[{Cimiano et~al.(2013)Cimiano, L\"{u}ker, Nagel, and
  Unger}]{cimiano2013exploiting}
Philipp Cimiano, Janna L\"{u}ker, David Nagel, and Christina Unger. 2013.
\newblock \href {http://www.aclweb.org/anthology/W13-2102} {Exploiting ontology
  lexica for generating natural language texts from rdf data}.
\newblock In \emph{Proceedings of the 14th European Workshop on Natural
  Language Generation}, pages 10--19, Sofia, Bulgaria. Association for
  Computational Linguistics.

\bibitem[{Colin et~al.(2016)Colin, Gardent, Mrabet, Narayan, and
  Perez-Beltrachini}]{colin2016webnlg}
Emilie Colin, Claire Gardent, Yassine Mrabet, Shashi Narayan, and Laura
  Perez-Beltrachini. 2016.
\newblock The webnlg challenge: Generating text from dbpedia data.
\newblock In \emph{Proceedings of the 9th International Natural Language
  Generation conference}, pages 163--167.

\bibitem[{Cucerzan(2007)}]{Cucerzan07}
Silviu Cucerzan. 2007.
\newblock Large-scale named entity disambiguation based on wikipedia data.
\newblock In \emph{Conference on Empirical Methods in Natural Language
  Processing-CoNLL}.

\bibitem[{Duma and Klein(2013)}]{duma2013generating}
Daniel Duma and Ewan Klein. 2013.
\newblock Generating natural language from linked data: Unsupervised template
  extraction.
\newblock In \emph{IWCS}, pages 83--94.

\bibitem[{Ell and Harth(2014)}]{ell2014language}
Basil Ell and Andreas Harth. 2014.
\newblock A language-independent method for the extraction of rdf verbalization
  templates.
\newblock In \emph{INLG}, pages 26--34.

\bibitem[{Ell et~al.(2011)Ell, Vrande{\v{c}}i{\'c}, and
  Simperl}]{ell2011labels}
Basil Ell, Denny Vrande{\v{c}}i{\'c}, and Elena Simperl. 2011.
\newblock Labels in the web of data.
\newblock \emph{ISWC}.

\bibitem[{van Erp et~al.(2016)van Erp, Mendes, Paulheim, Ilievski, Plu, Rizzo,
  and Waitelonis}]{vanErp2016}
Marieke van Erp, Pablo Mendes, Heiko Paulheim, Filip Ilievski, Julien Plu,
  Giuseppe Rizzo, and Joerg Waitelonis. 2016.
\newblock Evaluating entity linking: An analysis of current benchmark datasets
  and a roadmap for doing a better job.
\newblock In \emph{Proceedings of LREC}.

\bibitem[{Freitas et~al.(2010)Freitas, Carvalho, Gon{\c{c}}alo~Oliveira, Mota,
  and Santos}]{freitas2010second}
Cl{\'a}udia Freitas, Paula Carvalho, Hugo Gon{\c{c}}alo~Oliveira, Cristina
  Mota, and Diana Santos. 2010.
\newblock Second harem: advancing the state of the art of named entity
  recognition in portuguese.
\newblock In \emph{quot; In Nicoletta Calzolari; Khalid Choukri; Bente
  Maegaard; Joseph Mariani; Jan Odijk; Stelios Piperidis; Mike Rosner; Daniel
  Tapias (ed) Proceedings of the International Conference on Language Resources
  and Evaluation (LREC 2010)(Valletta 17-23 May de 2010) European Language
  Resources Association}. European Language Resources Association.

\bibitem[{Ganea et~al.(2016)Ganea, Ganea, Lucchi, Eickhoff, and
  Hofmann}]{Ganea:2016:PBM:2872427.2882988}
Octavian-Eugen Ganea, Marina Ganea, Aurelien Lucchi, Carsten Eickhoff, and
  Thomas Hofmann. 2016.
\newblock Probabilistic bag-of-hyperlinks model for entity linking.
\newblock In \emph{Proceedings of the 25th International Conference on World
  Wide Web}, WWW '16, pages 927--938, Republic and Canton of Geneva,
  Switzerland. International World Wide Web Conferences Steering Committee.

\bibitem[{Gardent et~al.(2017)Gardent, Shimorina, Narayan, and
  Perez-Beltrachini}]{gardent2017creating}
Claire Gardent, Anastasia Shimorina, Shashi Narayan, and Laura
  Perez-Beltrachini. 2017.
\newblock Creating training corpora for nlg micro-planning.
\newblock In \emph{Proceedings of ACL}.

\bibitem[{Hewlett et~al.(2005)Hewlett, Kalyanpur, Kolovski, and
  Halaschek-Wiener}]{hewlett2005effective}
Daniel Hewlett, Aditya Kalyanpur, Vladimir Kolovski, and Christian
  Halaschek-Wiener. 2005.
\newblock Effective nl paraphrasing of ontologies on the semantic web.
\newblock In \emph{Workshop on end-user semantic web interaction, 4th int.
  semantic web conference, galway, ireland}.

\bibitem[{Jha et~al.(2017)Jha, R{\"o}der, and Ngomo}]{eaglet_ESWC}
Kunal Jha, Michael R{\"o}der, and Axel-Cyrille~Ngonga Ngomo. 2017.
\newblock All that glitters is not gold--rule-based curation of reference
  datasets for named entity recognition and entity linking.
\newblock In \emph{ISWC}.

\bibitem[{Mendes et~al.(2011)Mendes, Jakob, Garcia-Silva, and
  Bizer}]{spotlight}
Pablo~N. Mendes, Max Jakob, Andres Garcia-Silva, and Christian Bizer. 2011.
\newblock {DBpedia Spotlight: Shedding Light on the Web of Documents}.
\newblock In \emph{7th International Conference on Semantic Systems
  (I-Semantics)}, pages 1--8.

\bibitem[{Milne and Witten(2008)}]{milne2008learning}
David Milne and Ian~H Witten. 2008.
\newblock Learning to link with wikipedia.
\newblock In \emph{ACM CIKM}.

\bibitem[{Mohammed et~al.(2016)Mohammed, Perez-Beltrachini, and
  Gardent}]{mohammed2016category}
Rania Mohammed, Laura Perez-Beltrachini, and Claire Gardent. 2016.
\newblock Category-driven content selection.
\newblock In \emph{Proceedings of the 9th International Natural Language
  Generation conference}, pages 94--98.

\bibitem[{Moro et~al.(2014)Moro, Cecconi, and Navigli}]{moro2014multilingual}
Andrea Moro, Francesco Cecconi, and Roberto Navigli. 2014.
\newblock Multilingual word sense disambiguation and entity linking for
  everybody.
\newblock In \emph{Proceedings of the 2014 International Conference on Posters
  \& Demonstrations Track-Volume 1272}, pages 25--28. CEUR-WS. org.

\bibitem[{Moussallem et~al.(2018)Moussallem, Ferreira, Zampieri, Cavalcanti,
  Xexeo, Neves, and Ngomo}]{LRECMoussallem2017}
Diego Moussallem, Thiago~Castro Ferreira, Marcos Zampieri, Maria~Claudia
  Cavalcanti, Geraldo Xexeo, Mariana Neves, and Axel-Cyrille~Ngonga Ngomo.
  2018.
\newblock {RDF2PT: Generating Brazilian Portuguese Texts from RDF Data}.
\newblock In \emph{LREC}.

\bibitem[{Moussallem et~al.(2017)Moussallem, Usbeck, R{\"o}der, and {Ngonga
  Ngomo}}]{Moussallem2017}
Diego Moussallem, Ricardo Usbeck, Michael R{\"o}der, and Axel-Cyrille {Ngonga
  Ngomo}. 2017.
\newblock {MAG: A Multilingual, Knowledge-base Agnostic and Deterministic
  Entity Linking Approach}.
\newblock In \emph{K-CAP: Knowledge Capture Conference}, page~8. ACM.

\bibitem[{Mrabet et~al.(2016)Mrabet, Vougiouklis, Kilicoglu, Gardent,
  Demner-Fushman, Hare, and Simperl}]{mrabet2016aligning}
Yassine Mrabet, Pavlos Vougiouklis, Halil Kilicoglu, Claire Gardent, Dina
  Demner-Fushman, Jonathon Hare, and Elena Simperl. 2016.
\newblock Aligning texts and knowledge bases with semantic sentence
  simplification.
\newblock \emph{WebNLG 2016}.

\bibitem[{{Ngonga Ngomo} et~al.(2013){Ngonga Ngomo}, B{\"u}hmann, Unger,
  Lehmann, and Gerber.}]{ngo+13a}
Axel-Cyrille {Ngonga Ngomo}, Lorenz B{\"u}hmann, Christina Unger, Jens Lehmann,
  and Daniel Gerber. 2013.
\newblock Sorry, i don't speak sparql --- translating sparql queries into
  natural language.
\newblock In \emph{Proceedings of WWW}, pages 977--988.

\bibitem[{{Ngonga Ngomo} and Schumacher(2009)}]{ngomo2009borderflow}
Axel-Cyrille {Ngonga Ngomo} and Frank Schumacher. 2009.
\newblock Borderflow: A local graph clustering algorithm for natural language
  processing.
\newblock In \emph{Computational Linguistics and Intelligent Text Processing},
  pages 547--558. Springer.

\bibitem[{Nuzzolese et~al.(2015)Nuzzolese, Gentile, Presutti, Gangemi,
  Garigliotti, and Navigli}]{okechallenge}
Andrea~Giovanni Nuzzolese, Anna~Lisa Gentile, Valentina Presutti, Aldo Gangemi,
  Dar{\'\i}o Garigliotti, and Roberto Navigli. 2015.
\newblock Open knowledge extraction challenge.
\newblock In \emph{Semantic Web Evaluation Challenge}.

\bibitem[{Oramas et~al.(2016)Oramas, Anke, Sordo, Saggion, and
  Serra}]{OramasASSS16}
Sergio Oramas, Luis~Espinosa Anke, Mohamed Sordo, Horacio Saggion, and Xavier
  Serra. 2016.
\newblock {ELMD:} an automatically generated entity linking gold standard
  dataset in the music domain.
\newblock In \emph{{LREC}}.

\bibitem[{Perez-Beltrachini et~al.(2016)Perez-Beltrachini, Sayed, and
  Gardent}]{perez2016building}
Laura Perez-Beltrachini, Rania Sayed, and Claire Gardent. 2016.
\newblock Building rdf content for data-to-text generation.
\newblock In \emph{COLING}, pages 1493--1502.

\bibitem[{Ratinov et~al.(2011)Ratinov, Roth, Downey, and Anderson}]{rat:rot}
Lev Ratinov, Dan Roth, Doug Downey, and Mike Anderson. 2011.
\newblock Local and global algorithms for disambiguation to wikipedia.
\newblock In \emph{Proceedings of the 49th Annual Meeting of the Association
  for Computational Linguistics: Human Language Technologies}, pages
  1375--1384.

\bibitem[{Rebholz-Schuhmann et~al.(2010)Rebholz-Schuhmann, Yepes, Van~Mulligen,
  Kang, Kors, Milward, Corbett, Buyko, Beisswanger, and
  Hahn}]{rebholz2010calbc}
Dietrich Rebholz-Schuhmann, Antonio Jos{\'e}~Jimeno Yepes, Erik~M Van~Mulligen,
  Ning Kang, Jan Kors, David Milward, Peter Corbett, Ekaterina Buyko, Elena
  Beisswanger, and Udo Hahn. 2010.
\newblock Calbc silver standard corpus.
\newblock \emph{Journal of bioinformatics and computational biology}.

\bibitem[{R{\"o}der et~al.(2014)R{\"o}der, Usbeck, Gerber, Hellmann, and
  Both}]{N3}
Michael R{\"o}der, Ricardo Usbeck, Daniel Gerber, Sebastian Hellmann, and
  Andreas Both. 2014.
\newblock {$\mbox{N}^3$ - A Collection of Datasets for Named Entity Recognition
  and Disambiguation in the NLP Interchange Format}.
\newblock In \emph{LREC}.

\bibitem[{Rosales{-}M{\'{e}}ndez et~al.(2018)Rosales{-}M{\'{e}}ndez, Hogan, and
  Poblete}]{voxeliswc2018}
Henry Rosales{-}M{\'{e}}ndez, Aidan Hogan, and Barbara Poblete. 2018.
\newblock Voxel: A benchmark dataset for multilingual entity linking.
\newblock In \emph{International Semantic Web Conference}. Springer.

\bibitem[{Schwitter et~al.(2004)Schwitter, Tilbrook
  et~al.}]{schwitter2004controlled}
Rolf Schwitter, Marc Tilbrook, et~al. 2004.
\newblock Controlled natural language meets the semantic web.
\newblock In \emph{Proceedings of the Australasian Language Technology
  Workshop}, volume~2, pages 55--62.

\bibitem[{Sleimi and Gardent(2016)}]{sleimi2016generating}
Amin Sleimi and Claire Gardent. 2016.
\newblock Generating paraphrases from dbpedia using deep learning.
\newblock \emph{WebNLG 2016}, page~54.

\bibitem[{Soto et~al.(2017)Soto, Gallardo, and Diz}]{soto2017adapting}
Alejandro~Ramos Soto, Julio~Janeiro Gallardo, and Alberto~Bugar{\'\i}n Diz.
  2017.
\newblock Adapting simplenlg to spanish.
\newblock In \emph{Proceedings of the 10th International Conference on Natural
  Language Generation}, pages 144--148.

\bibitem[{Staykova(2014)}]{staykova2014natural}
Kamenka Staykova. 2014.
\newblock Natural language generation and semantic technologies.
\newblock \emph{Cybernetics and Information Technologies}, 14(2):3--23.

\bibitem[{Steinmetz et~al.(2013)Steinmetz, Knuth, and Sack}]{yovisto}
Nadine Steinmetz, Magnus Knuth, and Harald Sack. 2013.
\newblock Statistical analyses of named entity disambiguation benchmarks.
\newblock In \emph{1st Workshop on NLP\&DBpedia 2013}, pages 91--102.

\bibitem[{Sun and Mellish(2006)}]{sun2006domain}
Xiantang Sun and Chris Mellish. 2006.
\newblock Domain independent sentence generation from rdf representations for
  the semantic web.
\newblock In \emph{Combined Workshop on Language-Enabled Educational Technology
  and Development and Evaluation of Robust Spoken Dialogue Systems, European
  Conference on AI, Riva del Garda, Italy}.

\bibitem[{{Tjong Kim Sang} and {De Meulder}(2003)}]{conll2003}
Erik~F. {Tjong Kim Sang} and Fien {De Meulder}. 2003.
\newblock Introduction to the conll-2003~{s}hared task: language-independent
  named entity recognition.
\newblock In \emph{Proceedings of the seventh conference on Natural language
  learning at HLT-NAACL 2003 - Volume 4}, pages 142--147.

\bibitem[{Tsatsaronis et~al.(2012)Tsatsaronis, Schroeder, Paliouras,
  Almirantis, Androutsopoulos, Gaussier, Gallinari, Artieres, Alvers, Zschunke,
  and {Ngonga Ngomo}}]{bioasq}
George Tsatsaronis, Michael Schroeder, Georgios Paliouras, Yannis Almirantis,
  Ion Androutsopoulos, Eric Gaussier, Patrick Gallinari, Thierry Artieres,
  {Michael R.} Alvers, Matthias Zschunke, and Axel-Cyrille {Ngonga Ngomo}.
  2012.
\newblock {BioASQ}: A challenge on large-scale biomedical semantic indexing and
  question answering.
\newblock In \emph{AAAI Information Retrieval and Knowledge Discovery in
  Biomedical Text}.

\bibitem[{Usbeck et~al.(2015)Usbeck, R\"{o}der, Ngonga~Ngomo, Baron, Both,
  Br\"{u}mmer, Ceccarelli, Cornolti, Cherix, Eickmann, Ferragina, Lemke, Moro,
  Navigli, Piccinno, Rizzo, Sack, Speck, Troncy, Waitelonis, and
  Wesemann}]{gerbil}
Ricardo Usbeck, Michael R\"{o}der, Axel-Cyrille Ngonga~Ngomo, Ciro Baron,
  Andreas Both, Martin Br\"{u}mmer, Diego Ceccarelli, Marco Cornolti, Didier
  Cherix, Bernd Eickmann, Paolo Ferragina, Christiane Lemke, Andrea Moro,
  Roberto Navigli, Francesco Piccinno, Giuseppe Rizzo, Harald Sack, Ren{\'e}
  Speck, Rapha\"{e}l Troncy, J\"{o}rg Waitelonis, and Lars Wesemann. 2015.
\newblock Gerbil: General entity annotator benchmarking framework.
\newblock In \emph{WWW '15}.

\bibitem[{Voyer et~al.(2010)Voyer, Nygaard, Fitzgerald, and
  Copperman}]{voyer2010}
Robert Voyer, Valerie Nygaard, Will Fitzgerald, and Hannah Copperman. 2010.
\newblock A hybrid model for annotating named entity training corpora.
\newblock In \emph{Proceedings of the 4th Linguistic Annotation Workshop}.

\end{thebibliography}
\bibliographystyle{acl_natbib.bst}

\end{document}